\RequirePackage{fix-cm}
\documentclass[twocolumn]{svjour3}          
\smartqed  
\usepackage{graphicx}
\usepackage[square]{natbib}
\usepackage{amsmath}
\usepackage{algorithmic}
\usepackage{algorithm}
\usepackage[pagebackref=true,breaklinks=true,letterpaper=true,colorlinks,bookmarks=false,citecolor=blue,linkcolor=blue]{hyperref}
\usepackage{makecell}
 
\begin{document}
\begin{sloppypar}
\title{Weakly-Supervised Semantic Segmentation by Iterative Affinity Learning
}


\author{Xiang Wang \and  Sifei Liu \and Huimin Ma$^\dagger$ \and Ming-Hsuan Yang
}


\institute{Xiang Wang  \at
              Tencent Research, Beijing, China \\
              Tsinghua University, Beijing, China \\
              \email{andyxwang@tencent.com}           
           \and
           Sifei Liu \at
           Nvidia, Santa Clara, CA, USA\\
           \email{sifeil@nvidia.com}    
           \and
           Huimin Ma \at
           University of Science and Technology Beijing, Beijing, China \\
           \email{mhmpub@ustb.edu.cn}    \\
           $\dagger$ corresponding author
           \and
           Ming-Hsuan Yang \at
           University of California at Merced, Merced, CA, USA\\
           \email{mhyang@ucmerced.edu}        
}

\date{Received: date / Accepted: date}

\maketitle

\begin{abstract}
Weakly-supervised semantic segmentation is a challenging task as no pixel-wise label information is provided for training. 
Recent methods have exploited classification networks to localize objects by selecting regions with strong response. 
While such response map provides sparse information, however, there exist strong pairwise relations between pixels in natural images, which can be utilized to propagate the sparse map to a much denser one.
In this paper, we propose an iterative algorithm to learn such pairwise relations, which consists of two branches, a unary segmentation network which learns the label probabilities for each pixel, and a pairwise affinity network which learns affinity matrix and refines the probability map generated from the unary network. 
The refined results by the pairwise network are then used as supervision to train the unary network, and the procedures are conducted iteratively to obtain better segmentation progressively.
To learn reliable pixel affinity without accurate annotation, we also propose to mine confident regions. 
We show that iteratively training this framework is equivalent to optimizing an energy function with convergence to a local minimum.
Experimental results on the PASCAL VOC 2012 and COCO datasets demonstrate that the proposed algorithm performs favorably against the state-of-the-art methods. 
\end{abstract}

\section{Introduction}
Semantic segmentation aims to predict a label for each pixel from a set of pre-defined object classes. 
With the advances of Deep Neural Networks (DNNs), significant progress has been made in semantic segmentation (\cite{long2015fully,zhao2017pyramid,chen2018deeplab,chen2017rethinking,zhou2019semantic}). 
However, fully-supervised methods require a large amount of pixel-wise annotations, which is time-consuming and expensive. 
To make semantic segmentation more practical, a number of weakly-supervised methods have been proposed in recent years based on partial information of each image, such as bounding boxes~(\cite{dai2015boxsup, khoreva2017simple}), scribbles~(\cite{lin2016scribblesup}), points~(\cite{bearman2016s}), and even class labels~(\cite{pathak2015constrained, wang2018weakly,ahn2018learning,huang2018weakly,wei2018revisiting}). 
In this paper, we present a weakly-supervised semantic segmentation algorithm based only on class labels of an image. 

\begin{figure}[t] 
\begin{center}
\includegraphics[width=1\linewidth,trim = 0mm 0mm 0mm 0mm, clip]{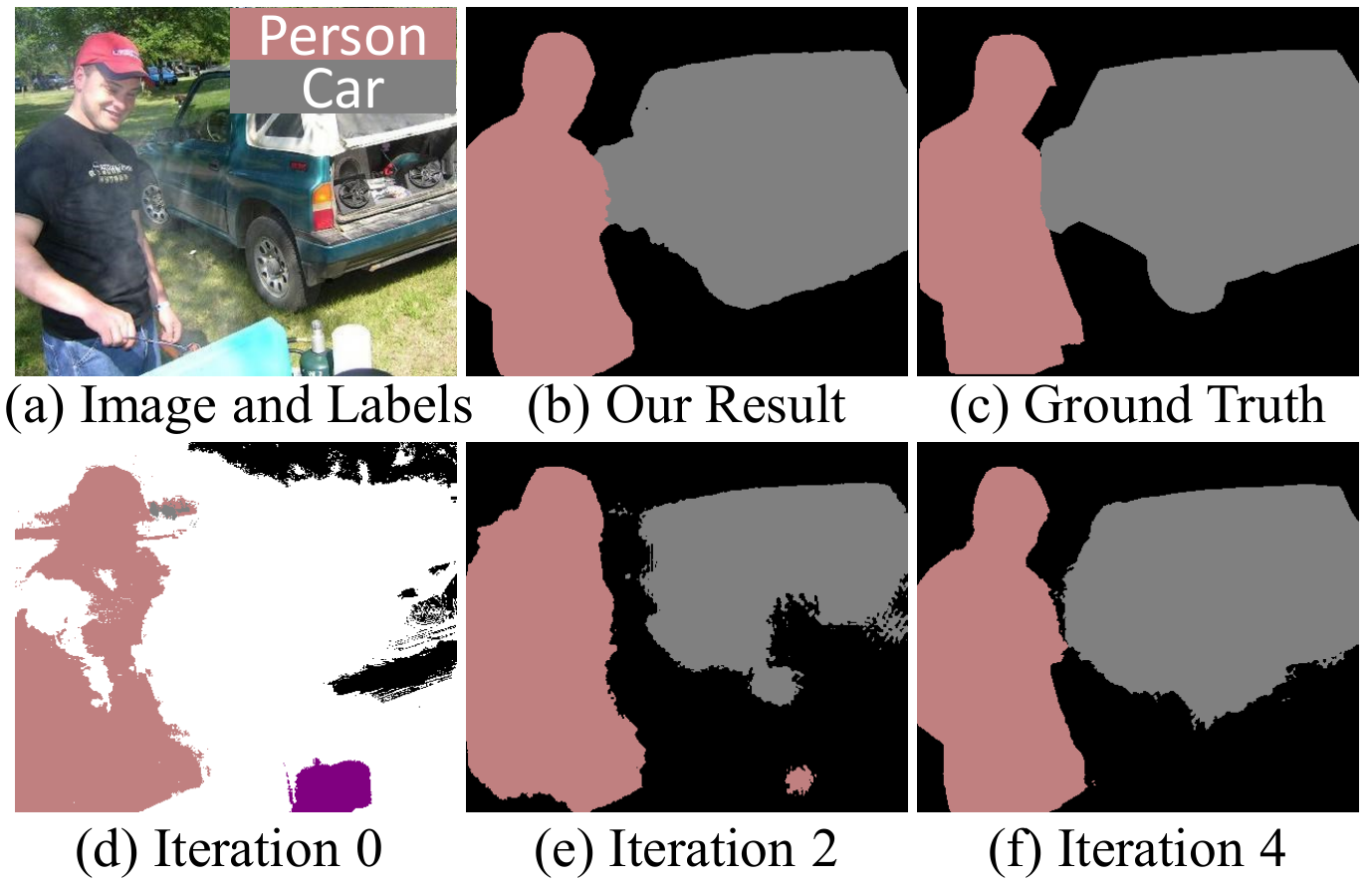}

\caption{Top row: Given training images and their class labels, our framework generates accurate segmentation results. Bottom row: By iteratively learning affinity, our framework progressively generates better segments for supervising the segmentation network. The seed regions generated by the CAM method~(\cite{zhou2016learning}) are shown in (d) where white color pixels denote image locations with unknown labels.}\label{fig-impressive}
\end{center}
\end{figure}

Weakly-supervised semantic segmentation based on class labels is challenging as no pixel in an image is annotated (\textit{i.e.}, an image is only annotated with class labels as shown in Figure~\ref{fig-impressive}(a)). 
Recently, the Class Activation Map (CAM) method~(\cite{zhou2016learning}) has been developed to generate discriminative object seed regions with classification networks. 
Since coarse response maps are generated~(Figure~\ref{fig-impressive}(d)), these regions cannot be directly used to train an accurate segmentation network. 
As data redundancy often exists in natural images~(\cite{kersten1987predictability}), significant statistical dependencies among pixels in images can be exploited. 
We can learn similarities or affinities from images, and propagate sparse and noisy labels of object regions to generate dense and accurate annotations.
With weak supervision, this is challenging as there are no accurate pixel-wise annotations and the region labels from the CAM method are noisy and sometimes inaccurate.
To address these issues, we mine confident regions from the coarse pixel labels and then learn pixel affinities from them to refine the coarse labels. 
Iteratively, we mine confident regions from the refined results and learn more robust affinities until convergence.

\begin{figure*}[t] 
\begin{center}
\includegraphics[width=0.95\linewidth,trim = 0mm 0mm 0mm 0mm, clip]{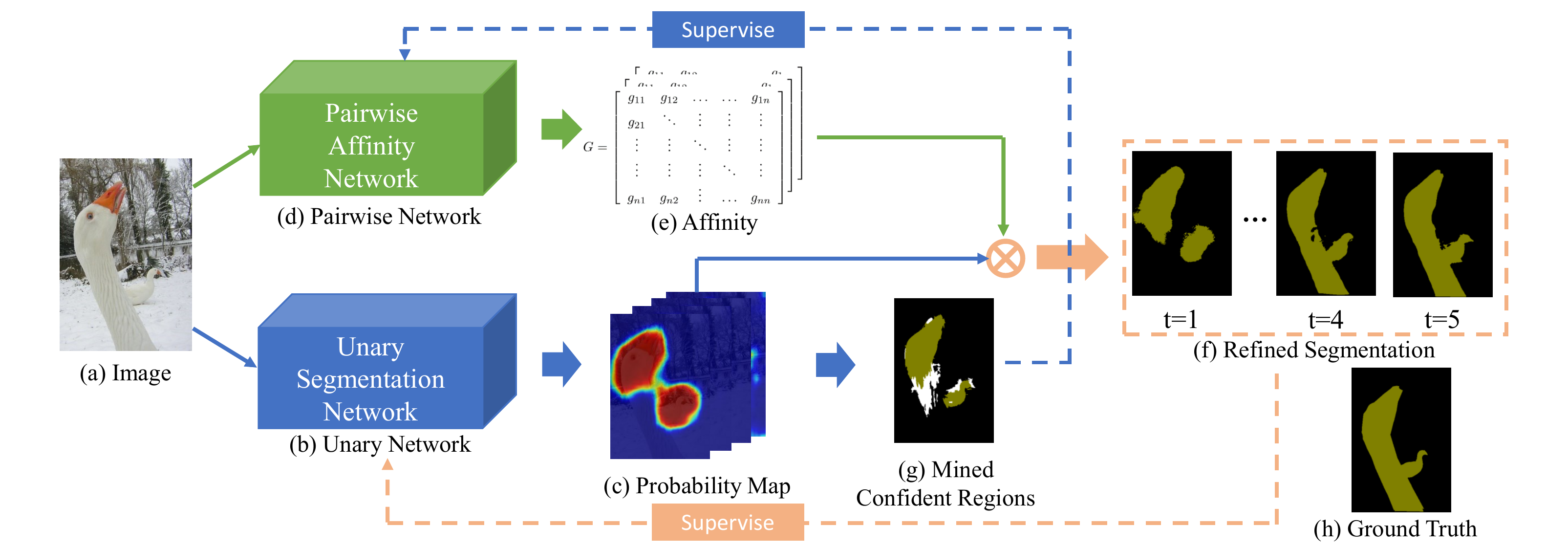}
\caption{
Illustration of the proposed framework. 
The framework consists of two branches: a (b) unary network which predicts
a (c) probability map of the input image, and a (d) pairwise network, which learns the (e) affinity matrix from the (g) mined confident regions. The learned affinities are then applied to the probability map from the unary network to generate the (f)
refined segmentation. 
The refined results are then used as supervision signals to retrain the unary network. 
These procedures are conducted iteratively to learn more robust affinity progressively and produce more accurate segmentation. 
}
\label{fig-framework}
\end{center}
\vspace{-3mm}
\end{figure*}

In this paper, we propose an iterative affinity learning framework, which consists of two major branches (see Figure~\ref{fig-framework}): 
a unary segmentation network which learns the pixel-wise probability of semantic categories from produced labels, and a pairwise network which refines the current labels by learning the affinity matrix and propagating the labels.
The refined results by the pairwise network provide better ``ground truth" to retrain the unary segmentation network in the next iteration. 
The above procedures are conducted iteratively until convergence to obtain better segmentation progressively.
Figure~\ref{fig-impressive} shows one example.
Given training images and the class labels, the proposed framework can generate accurate semantic segmentation results.
This is achieved by the iterative optimization strategy which learns reliable affinity and generates better masks for supervising the segmentation network.

The key ingredient of our framework is learning affinities between pixels, which determines the amount of improvements achieved at each iteration. 
However, under weak supervision, we do not have accurate annotations to learn pixel affinities.
To address this issue, we propose to mine confident regions from the output results of the unary network, and then use them to supervise the pairwise affinity network.
Our motivation is that, to learn the affinity, we only need to know some pixel samples, which indicate the pixels belonging to the same (their pixel affinity should be high) or different classes (their pixel affinity should be low).
Even with a small amount of pixel samples, we are able to learn segmentation by propagating and mining more labels via learning the affinity.
We also show that iteratively training the proposed framework is equivalent to optimizing an energy function with an EM-like approach. 
Furthermore, we show that this process always converges to a local minimum due to that the energy function is differentiable with respect to both the output labels and the network parameters.

The main contributions of this work are summarized as follows:

\begin{itemize}
\item We present an iterative affinity learning framework to progressively generate better segmentation, and show that it is equivalent to optimizing an energy loss function. 
We show that it always converges to a local minimum.

\item We propose a method to learn reliable affinity from inaccurate annotations by mining confident regions.

\item We demonstrate that the proposed weakly-supervised semantic segmentation algorithm performs favorably against the state-of-the-art methods on the PASCAL VOC 2012 and COCO datasets.

\end{itemize}

\section{Related Work}
In this section, we discuss related methods for weakly-supervised semantic segmentation and learning affinity for segmentation.
\subsection{Weakly-Supervised Semantic Segmentation}
Weakly-supervised semantic segmentation based on class labels has drawn much attention in recent years due to low annotation costs. 
Early methods~(\cite{pathak2014fully, pathak2015constrained, pinheiro2015image}) mainly formulate this problem as a multi-instance learning (MIL) problem.  
\cite{pathak2014fully} propose to add a max-pooling layer on top of FCN~(\cite{long2015fully}) and design a multi-class MIL loss for training the network. 
Based on this framework, several methods have been developed~(\cite{pathak2015constrained, pinheiro2015image}).
\cite{pathak2015constrained} add several constraints on foreground and suppression schemes to the MIL framework for weakly-supervised semantic segmentation. 
\cite{pinheiro2015image} replace the max-pooling layer of the MIL framework with a new \textit{Log-Sum-Exp} layer which can consider more information of the feature layers.

Recent methods~(\cite{kolesnikov2016seed, wei2017object,wang2018weakly, ahn2018learning,huang2018weakly,wei2018revisiting}) tackle weakly-supervised semantic segmentation by a two-stage procedure, which first generates initial object labels with class activation maps~(\cite{zhou2016learning}), and then trains segmentation networks based on the response maps.
\cite{kolesnikov2016seed} present an end-to-end framework with three modules (seed, expand and constrain) as loss functions, and the class activation maps are used as supervisory signals. 
A number of methods are developed to expand object regions based on the class activation maps.
\cite{wei2017object} propose to progressively erase most significant regions in the activation maps and then generate more regions. 
These regions are then used as ground truth to train a segmentation network. 
\cite{wang2018weakly} develop a bottom-up and top-down framework which iteratively mines common object features to expand initial object regions from the class activation maps.  
\cite{ahn2018learning} learn the pixel affinity from the activation maps and then apply the random walk method to refine them. 
\cite{huang2018weakly} design a deep seeded region growing algorithm which improves the seed regions to supervise the network.

Among the above-mentioned approaches, \cite{wei2017object} and~\cite{wang2018weakly} also use iterative strategies to refine segmentation results.  
However, the method of~\cite{wei2017object} heavily relies on the CAM network to progressively produce the most significant regions in the remaining images. 
Consequently, less discriminative object regions are usually missing. 
In addition, this method is not able to suppress noisy regions well. 
\cite{wang2018weakly} expand object regions by mining common object features. 
However, common features are only learned from each superpixel region and the pixel-wise context information is not exploited.
In contrast, our method can learn and propagate pixel-wise affinities to achieve better segmentation results.
We note \cite{ahn2018learning} also use the pixel affinities to refine segmentation results. 
However, the affinities are only learned from the coarse response map of the CAM method.
In the proposed framework, the pixel affinities are iteratively optimized, which are more reliable and lead to better segmentation results.  

\subsection{Learning Pixel Affinity for Segmentation}
An affinity matrix measures the similarities between pixels and has been widely used in object segmentation.
Some early methods directly define similarity functions to compute affinity matrices.
%
\cite{hagen1992new} propose a spectral methods for ratio cut~(\cite{wei1989towards}) which captures both min-cut and equipartition to locate natural clusters.
\cite{shi2000normalized} formulate image segmentation as a graph partitioning problem and present the normalized cut algorithm. 
This algorithm considers both the dissimilarity between different groups and the similarity within the same group.
%

In recent years, with the advances of DNNs, numerous algorithms have been proposed to learn the affinity 
end-to-end with deep networks~(\cite{liu2017learning, maire2016affinity,bertasius2017convolutional}).
\cite{maire2016affinity} present the affinity CNN which directly learns an affinity matrix to model pairwise relations for figure and ground embedding.  
\cite{liu2017learning} design the spatial propagation network (SPN) which directly learns pixel affinities and a spatial linear propagation module. 
The SPN takes images and coarse masks as input and learns pixel affinities end-to-end to refine the coarse masks.
\cite{bertasius2017convolutional} develop a random walk layer on top of the semantic segmentation network to learn the pixel affinities. 
%
%

These methods all learn the pixel affinities under full supervision to refine segmentation results. 
In contrast, our method aims to learn pixel affinities to refine object regions from coarse and inaccurate labels without pixel-wise annotations.
To address this challenging problem, we propose an iterative optimization framework which progressively mines confident regions for learning reliable affinity and generates better segmentation results. 

\section{Proposed Algorithm}
We solve the weakly-supervised semantic segmentation problem with an iterative optimization algorithm which progressively learns robust pixel affinities and propagates label information for accurate results. 
We present an EM approach that alternatively learns the network parameters for both unary segmentation and pairwise affinity networks, and maximizes the likelihood of the ``ground-truth" labels.
This is different from the fully-supervised approaches where the supervision requires the ground-truth labels.

\subsection{Formulation}
\label{sec-formulation}
Let $x$ denote an image. 
The proposed framework consists of two major branches (Figure~\ref{fig-framework}): (a) a unary network $F = f(x, W_f)$  parameterized by $W_f$ that learns the label probability with respect to each pixel in $x$, and (b) a pairwise network $G=g(x, W_g)$ that learns the pixel affinities, where $G\in {\cal{R}}^{N\times N}$, $N$ is the number of pixels, and $W_g$ is the parameter of the pairwise network. 
In addition, we denote $\alpha$ as the hidden state of the output labels.
%
We use the subscript $t$ to denote the $t^{th}$ step in the iterative process. 

We represent each image as an undirected weighted graph $\mathcal{G}={(V, E)}$, with the vertex set $V=\{v_1, \ldots, v_N\}$, where each edge between 
$v_i$ and $v_j$ has a weight $w_{ij}$.
The adjacency matrix is  $W=(w_{ij})_{i,j=1,\dots, N}$.
The degree matrix $D$ is a diagonal matrix with the degrees $d_1, \dots, d_N$ as elements, where $d_i=\sum\limits_{j=1}^{N}w_{ij}$.
The semantic segmentation problem is then to minimize the following energy loss function:
\begin{equation}\label{mainloss}
\alpha^* = \arg\min\limits_{\alpha}J(\alpha, W_f, W_g) =\arg\min\limits_{\alpha} \alpha^\top L \alpha,
\end{equation}
where $L = D - W$ is the Laplacian matrix, and 
\begin{equation}\label{equ-alphaloss}
\begin{split}
\alpha^\top L \alpha & =  \alpha^\top (D-W) \alpha \\
& = \sum\limits_{i=1}^N d_i \alpha_i^2-\sum\limits_{i,j=1}^N \alpha_i\alpha_j w_{i,j} \\
& = \frac{1}{2}\left( \sum\limits_{i=1}^N d_i \alpha_i^2 - 2\sum\limits_{i,j=1}^N \alpha_i\alpha_j w_{i,j} +  \sum\limits_{j=1}^N d_j \alpha_j^2\right)\\
& = \frac{1}{2}\sum\limits_{i,j=1}^N w_{i,j}(\alpha_i-\alpha_j)^2.
\end{split}
\end{equation}
That is, to minimize the loss function~(\ref{equ-alphaloss}) is to enforce pixels with high similarities (their affinity $w_{i,j}$ is high) to have similar labels.
This allows us to propagate label information for accurate results.

Instead of designing similarity metric and solve $\alpha$ as an optimization problem~(\cite{Levin2008ACS}), we propose an iterative learning method to refine the probability map and the networks via an EM formulation.
We denote $G=I-D+W=I-L$ as the affinity transformation matrix~(\cite{liu2017learning}), which is learnable by the pairwise network $g(x, W_g)$, and $\alpha^u$,  
$\alpha^p$ as the output of the unary network and the pairwise network, respectively.
The EM procedure are as follows:

\begin{itemize}
\item \textbf{Initialization}: We train the unary network and the pairwise network with object seeds $Y_0$ from class activation maps~(\cite{zhou2016learning}) to obtain the initial parameter $\{W_f, W_g\}_0$, the unary response map $\alpha_0^u$ (Figure \ref{fig-framework}(c)).
%
\item \textbf{E-step}: We refine the unary probability by minimizing $J_t$ w.r.t $\alpha_t^u$ given $W_f, W_g$, where:
\begin{equation}\label{Estep}
\frac{\partial{J}}{\partial{\alpha_t^u}} = L_t\alpha_t^u=(I-G_t)\alpha_t^u,
\end{equation}
and compute the refined map as $\alpha^p_{t}=\alpha_t^u-\Delta{\alpha_t^u}$ (i.e., $\alpha^p_{t}$ is the output of the pairwise network in step $t$).  
From \eqref{Estep} we have $\alpha^p_{t}=G_t\alpha_t^u$, where the corresponding network implementation is described in Section~\ref{sec-pnet}.
\item \textbf{M-step}: In this step, we minimize $J_t$ by learning both $W_f$ and $W_g$, through training the network $f_{t+1}(x, W_f)$ and $g_{t+1}(x, W_g)$ with the supervision signal extracted from $\alpha_{t}^p$ (Section~\ref{sec-unarynet}, \ref{sec-pnet}, \ref{sec-miningconf}).
\end{itemize}

It is straightforward to show the above procedures always converge to a local minimum, due to that 
$J$ is differentiable with respect to both $\alpha$ and the network parameters. 
However, to validate the M-step, we need to validate the link between the E-step and  M-step, \textit{i.e.}, 
how we use the network response from step $t$ to train $f_{t+1}(x, W_f)$ and $g_{t+1}(x, W_g)$ to minimize the energy function~(\ref{mainloss}).

For the unary network, in the $t+1$ step, it uses segmentation results of $\alpha^p_{t}$ as supervision to generate label probability $\alpha^u_{t+1}$.
For training the pairwise network in the $t+1$ step, we consider the softmax cross-entropy loss function with $\alpha^u_{t+1}$ as supervision:
\begin{equation}
H(\alpha^p_{t+1}) = -{\alpha^u_{t+1}}^\top\log\alpha^p_{t+1}.
\end{equation}
Since $\log(\cdot)$ is a monotonic increasing and convex function, optimizing $\alpha^p_{t+1}$ is to learn $G_{t+1}$ and $\alpha^p_{t+1} = G_{t+1}\alpha_{t+1}^u$. 
Therefore, minimizing $H$ is equivalent to minimize $-{\alpha^u_{t+1}}^\top\alpha^p_{t+1} = -{\alpha^u_{t+1}}^\top G_{t+1} \alpha_{t+1}^u= -{\alpha^u_{t+1}}^\top (I -L_{t+1}) \alpha_{t+1}^u$.
As the first term $-{\alpha^u_{t+1}}^\top\alpha_{t+1}^u$ is a constant, to optimize $\alpha_{t+1}^p$ is to minimize the second term:
\begin{equation}\label{proof}
L_{t+1}=\arg\min\limits_{L_{t+1}^*}{\alpha^u_{t+1}}^\top L_{t+1} \alpha_{t+1}^u,
\end{equation}
which is consistent with (\ref{mainloss}).
By using $\alpha^p_t$ as supervision to train the unary network and using ${\alpha^u_{t+1}}$ to supervise the pairwise network with the softmax cross-entropy loss, it is equivalent to minimize the original energy loss function. 
Namely, the objective of the M-step is to minimize the energy loss function.

However, in the stage of the pairwise network, if we use $\alpha_{t+1}^u$ to supervise the pairwise network to learn affinity and then refine itself, there is no information gain over iterations and the optimization will come to convergence to a relative low performance with very few steps (Section \ref{sec-aba-mineconf}). 
To obtain more accurate supervision in each step, we propose to mine confident regions from the output of the unary network $\alpha^u_{t+1}$.
These confident regions contain pixels belonging to object regions with high precision from which we can learn reliable affinity matrices (Section~\ref{sec-miningconf}).
We denote it as $Y_{t+1}$, and expect it to have lower energy (Section~\ref{sec-aba-mineconf}):
\begin{equation}\label{ytplus1}
    Y_{t+1}^\top L_{t+1}\alpha^u_{t+1} \le {\alpha^u_{t+1}}^\top L_{t+1}\alpha^u_{t+1}.
\end{equation}
With mining confident regions, our algorithm  converges to a lower energy and obtains better segmentation results.

The proposed EM procedures are summarized in Algorithm~\ref{algo}.

\begin{algorithm}[t]
    \hyphenation{object}
    \caption{Procedures of the proposed approach} 
    \hspace*{0.01in} {\bf Input:} \\
    \hspace*{0.15in} Generate object seeds from CAM, set it as $Y_0$. \\
    \hspace*{0.15in} Training images $x$.\\
    \hspace*{0.01in} {\bf Initialize:} \\
    \hspace*{0.15in} Train networks $f_{0}(x, W_f)$ and $g_{0}(x, W_g)$ with $Y_0$ to obtain the parameters $\{W_f, W_g\}_0$, the affinity matrix $G_0$ and the output of the unary network $\alpha^u_0$. 
    \begin{algorithmic}[1]
        \REQUIRE 
        \STATE Propagate $\alpha^u_t$ with $G_t$: $\alpha^p_{t}=G_t\alpha^u_t$ (Section~\ref{sec-pnet}).
        \ENSURE 
        \STATE Train $f_{t+1}(x, W_f)$ with $\alpha^p_{t}$ as supervision to obtain $\alpha^u_{t+1}$ (Section~\ref{sec-unarynet}).
        \STATE Mine confident regions $Y_{t+1}$ from the output of $f_{t+1}(x, W_f)$ (Section~\ref{sec-miningconf}).
        \STATE Train $g_{t+1}(x, W_g)$ with $Y_{t+1}$ as supervision to obtain $G_{t+1}$ (Section~\ref{sec-pnet}).
    \end{algorithmic}
    \label{algo}
\end{algorithm}

\subsection{Network Architecture and Training}
Figure~\ref{fig-framework} shows the architecture of the proposed framework. 
The framework consists of two major branches, a unary network $F = f(x, W_f)$ that learns the label probability of each pixel, and a pairwise network $G=g(x, W_g)$ that learns the affinity. 
The learned affinities are applied to the output probability map of the unary network to refine it and obtain better segmentation results. 

\subsubsection{Unary Network}\label{sec-unarynet}
The unary network aims to generate a probability map given a coarse segmentation mask. 
In this work, we use the  DeepLab~(\cite{chen2018deeplab}) model as the unary segmentation network.
To initialize the framework, we first generate object seed regions using the CAM method~(\cite{zhou2016learning}) in a way similar to~(\cite{ahn2018learning}).
The CAM method generates object regions for all classes, including background, and pixels with weak response are labelled as unknown, as shown in Figure~\ref{fig-impressive}(d).
We then use them as pseudo ground truth to train the unary segmentation network. 
The training process is the same as fully-supervised methods with a softmax loss as the objective function.
With this segmentation network, probability maps are generated for all classes.

\subsubsection{Pairwise Network}\label{sec-pnet}
The pairwise network aims to learn the pixel affinities from object regions and then applies to the probability maps to refine the segmentation results. 
In this work, we use the Spatial Propagation Network (SPN)~(\cite{liu2017learning}) to learn pairwise affinities. 
The SPN learns the affinity transformation matrix from an image $x$ to refine the coarse probability maps $\alpha^u_t$ and generates better segmentation $\alpha^p_{t}$.
It is an end-to-end framework which simultaneously learns the affinity transformation matrix $G$ and outputs the refined segmentation $\alpha^p_{t}=G_{t}\alpha^u_t$.
When learning the affinity, we raster scan the pixels from four directions: left-to-right, top-to-bottom, and vice versa. 
Since we use all three RGB image channels, we learn 12 affinity matrices.
More details regarding the spatial propagation network can be found in~\cite{liu2017learning}.

\begin{figure}[t]
\begin{center}
\includegraphics[width=1\linewidth,trim = 0mm 0mm 0mm 0mm, clip]{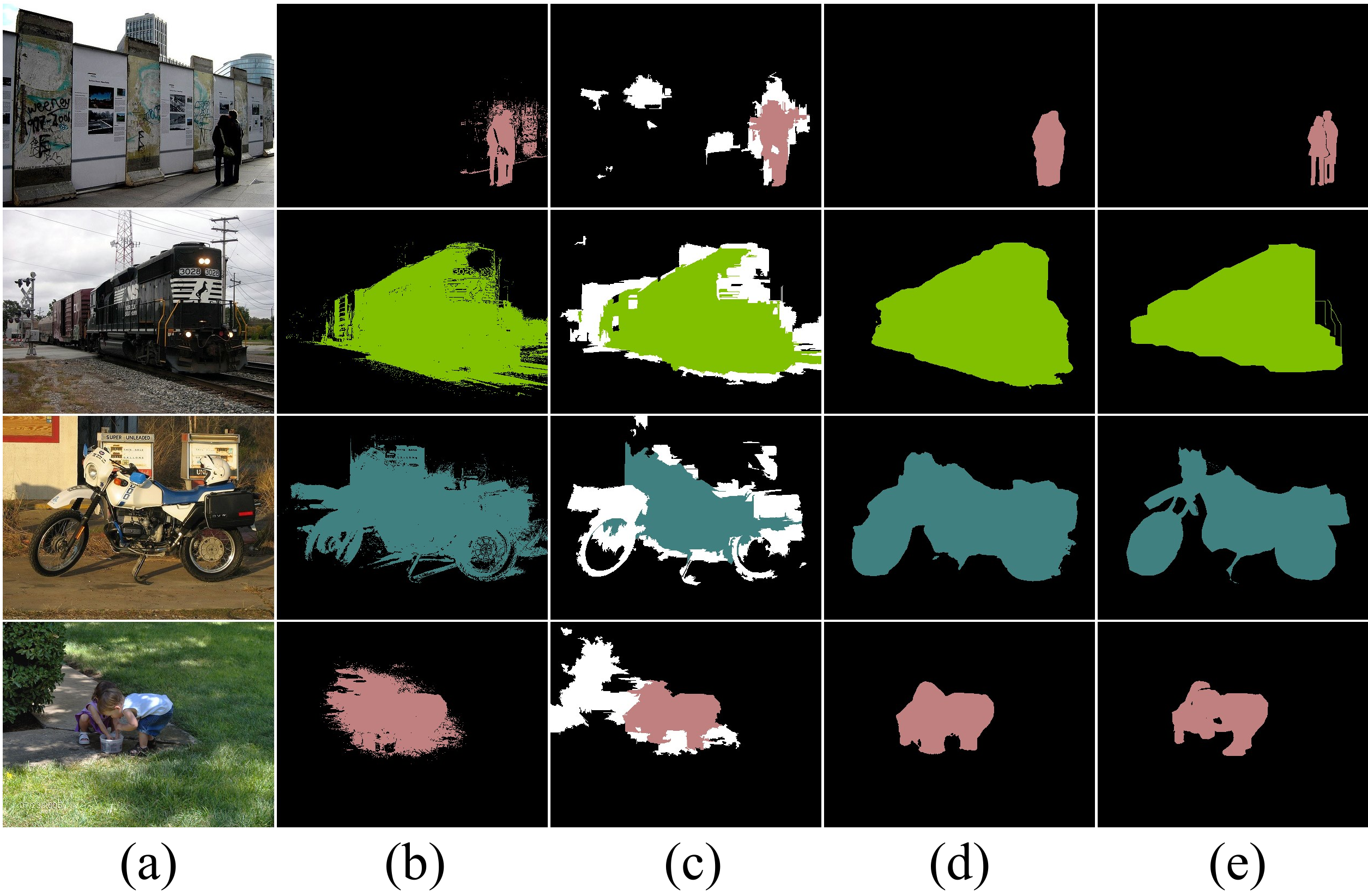}
\end{center}
\caption{Some examples of mining confident regions from segmentation results of the unary network: (a) images, (b) segmentation results of the unary network, (c) mined confident regions, (d) refined results of the pairwise affinity network, (e) ground truth. 
White color pixels denote image locations with unknown labels.}
\label{fig-mineconf}
\end{figure}

\begin{figure*}
\begin{center}
\includegraphics[width=1\linewidth,trim = 0mm 0mm 0mm 0mm, clip]{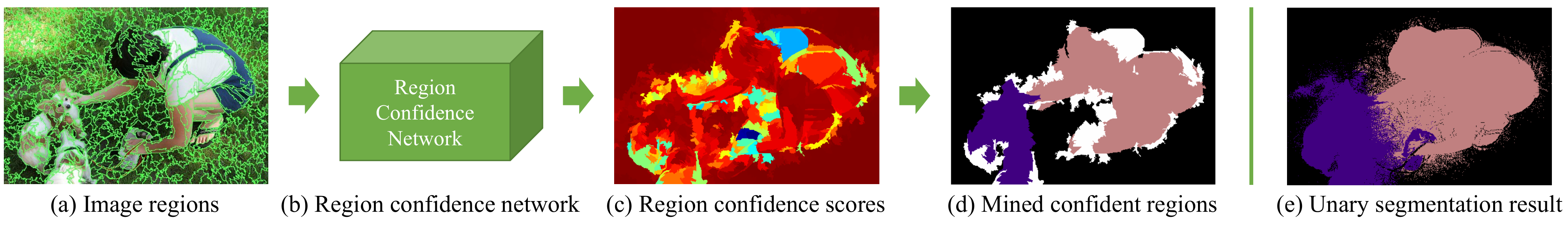}
\end{center}
\caption{Illustration of mining confident regions. Given an (a) input image, we segment it into superpixel regions, and apply the learned (b) region confidence network to predict object classes, and generate (c) confidence scores for all regions.
By selecting regions with high confidence score, we can obtain the (d) mined confident regions. White color pixels denote image regions with unknown labels. 
Compared with the (e) unary segmentation result, the noisy regions are mostly removed and some regions are corrected.
Thus, regions with high precision are extracted and used to better supervise the pairwise network. }
\label{fig-mineconfpipeline}
\end{figure*}

The spatial propagation network has been shown to perform well in pixel labelling under full supervision~(\cite{liu2017learning}).
However, under weak supervision, it is challenging to train the pairwise affinity network 
as no pixel-wise annotations are provided.
A straight-forward approach is to use the segmentation result at time $t$ as ground truth to supervise the pairwise affinity network. 
However, as the segmentation results are not accurate, the affinity matrix cannot be learned well. 
To address this issue, we first mine confident regions from the segmentation results and then learn the affinity matrix from the mined confident regions. 
If we can obtain some confident regions which have high precision to identify the class each region belongs to, we know that pixels within same class should have high affinities and pixels of different classes should have low affinities.
Thus, we can also learn reliable affinity matrices. 
For regions with low confidence scores, we mark them with unknown labels when training the pairwise affinity network.
Namely, when computing the softmax loss function, these regions are ignored.
The details of mining confident regions are introduced in Section~\ref{sec-miningconf}. 
We denote the mined confident regions as $Y_{t}$.
To train the pairwise affinity network, we utilize the softmax loss:
\begin{equation}
\mathcal{L}_a = -Y_{t}^\top\log{\alpha^p_{t}}.
\end{equation}
To learn accurate affinity matrices, we also introduce a \textit{region smoothness loss}. 
Our motivation is that a good affinity matrix should have similar values for pixels in the same object regions such that the refined results can be smooth and have clear boundaries.
To achieve this, we average the learned affinity matrix $G$ within each superpixel region and denote it as $G_s$. %
The objective function is then to minimize the difference between $G$ and $G_s$:
\begin{equation}
\mathcal{L}_s = \|G-G_s\|_2^2.
\end{equation}

\subsection{Mining Confident Regions}\label{sec-miningconf}
The key issue in our framework is how to learn reliable affinity matrices without accurate annotations. 
Our solution is to mine confident regions. 
We expect these confident regions contain pixels belonging to object regions with high precision from which we can learn reliable affinity matrices. 

Our method is based on statistical learning. 
The object regions generated by the unary network contain noisy results.
For example, some background pixels may be recognized as parts of an object, as shown in Figure~\ref{fig-mineconf}(b). 
We can learn a confidence score for each region with the segmentation results of the unary network as training samples. 
For a region with a certain class label, its initial confidence score is set as $1$ for this object, and $0$ for other objects.
By training a multi-class classification network with these regions, each region is assigned with a new confidence score.
For region pixels that have high similarity to one object, they will receive high confidence scores.
For region pixels different from one object (\textit{i.e.}, noisy regions), they will receive low confidence scores. 
With this procedure, we can remove noisy regions and select confident regions with high confidence scores. 
In this paper, we set the threshold as $0.7$ based on our empirical observations.
Some examples are shown in Figure~\ref{fig-mineconf}(c). 
With these confident regions, we can learn reliable affinity matrices from them and thus generate more accurate segmentation results~(Figure~\ref{fig-mineconf}(d)).

We first segment images into superpixel regions~(\cite{felzenszwalb2004efficient}) $\mathcal{S}=\{S_{i,j}\}$, where $S_{i,j}$ denotes the $j$-th superpixel in the $i$-th image.
For each region, its class label is obtained from the segmentation results.
If more than 80\% pixels of a superpixel is marked with a certain class $c$ in the segmentation results, then this superpixel is considered as a sample of class $c$. 
This scheme is formulated with the one-hot encoding, namely, $L_{i,j} = [0,\ldots,1,\ldots,0]$, where $L_{i,j}(c)=1$, 
$L_{i,j}(k)=0~(k=0,\ldots,C, k\neq c)$, and 
$C$ is the number of classes.
With the superpixel regions and corresponding labels $\mathcal{D}=\{S, L\}_{i,j}$, we can train a region classification network $f_c^m$ parameterized by $\theta_m$ to obtain a confidence score for each region with the cross-entropy loss function:
\begin{equation}\label{mconfloss}
\mathcal{L}_m = - \sum\limits_{i,j,c}L_{i,j}(c)\log{f_c^m(S_{i,j}|\theta_m)}.
\end{equation}

We train the region confidence network with the architecture proposed by~(\cite{wang2018edge}) which is a variant of the fast R-CNN model with a mask pooling scheme.
Similar to recent weakly-supervised learning methods~(\cite{pathak2015constrained, kolesnikov2016seed, ahn2018learning, huang2018weakly}), we initialize this network with the weights of a pre-trained model based on the ImageNet.
The model is trained with  $\mathcal{D}=\{S, L\}_{i,j}$ using (\ref{mconfloss}) as the loss function, where the superpixel region $S$ is the input and the corresponding class label $L$ is the supervisory signal.
With this region confidence network, we extract features of all superpixel regions of an image in one forward pass, and then recognize their classes. 
Figure~\ref{fig-mineconfpipeline} shows the process of mining confident regions. 
With the trained region confidence network, we can re-predict each superpixel region of images, and obtain confidence scores for all regions~(Figure~\ref{fig-mineconfpipeline}(c)).
To extract regions with high precision for learning reliable affinities, we select regions with high confidence scores (e.g., $>0.7$ in this work), and leave others as unknown~(Figure~\ref{fig-mineconfpipeline}(d)). 
Namely, we do not use unknown regions for training the pairwise affinity network. 

\begin{figure*}[!h] 
\begin{center}
\begin{tabular}{c} 
\hspace{-4mm}
\includegraphics[width=1\linewidth,trim = 0mm 0mm 0mm 0mm, clip]{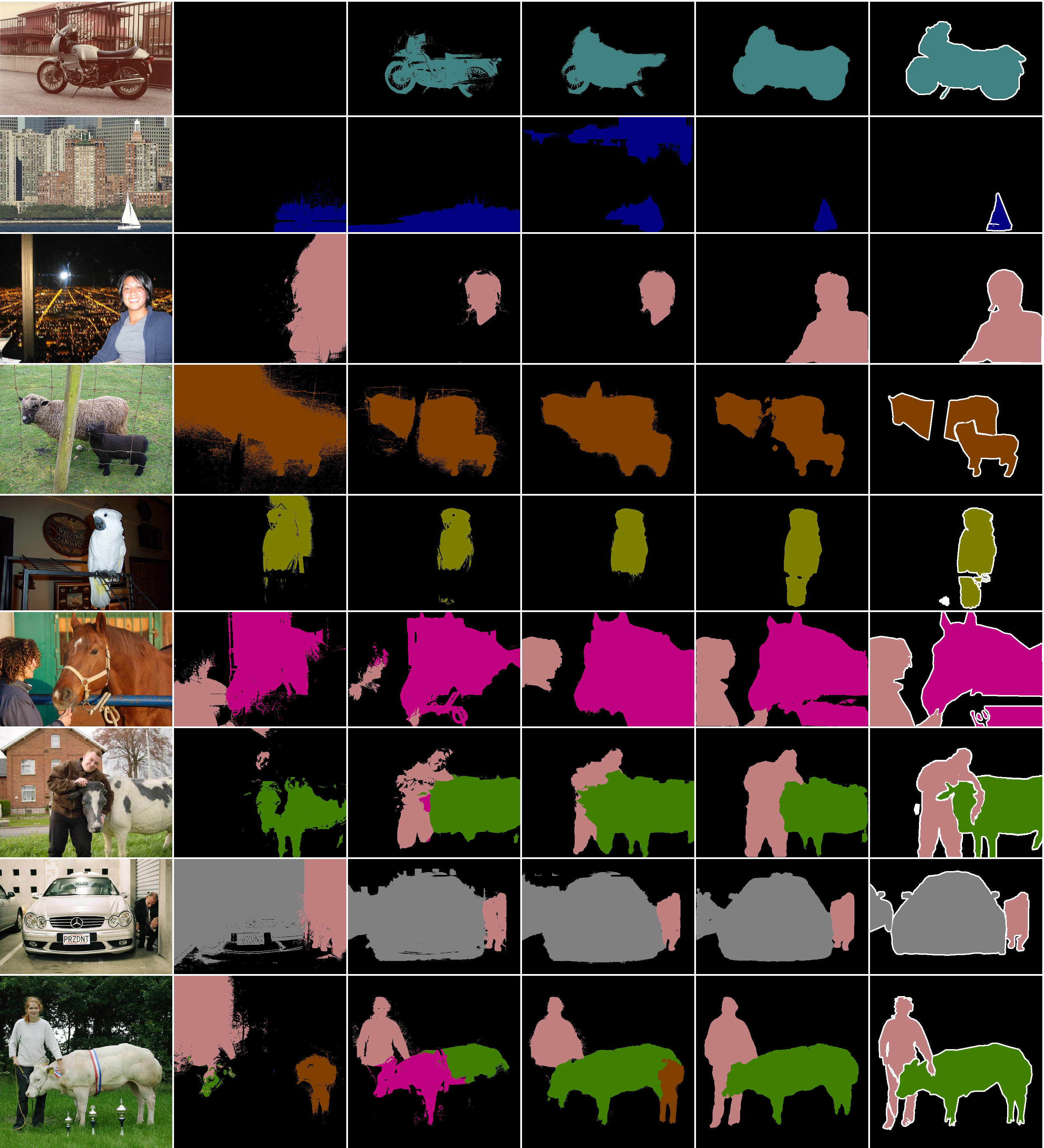}\\
\begin{minipage}[b]{0.16\linewidth}
  \centering
  \centerline{\footnotesize{Image}}\medskip
\end{minipage}
\begin{minipage}[b]{0.16\linewidth}
  \centering
  \centerline{\footnotesize{CCNN}}\medskip
\end{minipage}
\begin{minipage}[b]{0.16\linewidth}
  \centering
  \centerline{\footnotesize{SEC}}\medskip
\end{minipage}
\begin{minipage}[b]{0.16\linewidth}
  \centering
  \centerline{\footnotesize{MCOF}}\medskip
\end{minipage}
\begin{minipage}[b]{0.16\linewidth}
  \centering
  \centerline{\footnotesize{Ours}}\medskip
\end{minipage}
\begin{minipage}[b]{0.16\linewidth}
  \centering
  \centerline{\footnotesize{GT}}\medskip
\end{minipage}
\hfill
\end{tabular}
\caption{Visual comparisons with the state-of-the-art methods on the PASCAL VOC 2012 \textit{val} set.}
\label{fig-visres}
\end{center}
\end{figure*}

\section{Experimental Results}
\subsection{Settings}
We evaluate the proposed method on the PASCAL VOC 2012~(\cite{everingham2010pascal}) and  COCO~(\cite{lin2014microsoft}) datasets. 
The PASCAL VOC 2012 dataset contains $20$ object classes 
and $1$ background class with
1464 training images, 1449 validation images, and 1456 testing images. 
%
Same as the recent work~(\cite{wei2017object,ahn2018learning,huang2018weakly,wei2018revisiting,wang2018weakly}), we use the augmented set with 10582 images from~(\cite{hariharan2011semantic}) for training. 
For the COCO dataset, it contains more complex scenes and more classes ($80$ classes plus $1$ background class) with 80k images for training and 40k images for validation.
We iteratively train our framework on the training set using only class labels.
For inference, we forward the input images to the trained networks in the last iteration to obtain segmentation results, and the process is still efficient.
We evaluate the proposed algorithm against the state-of-the-art methods using the mean intersection-over-union (mIoU) metric. 

\subsection{Training Process}
The CAM Network is trained with the PyTorch framework and other models are trained with the Caffe package~(\cite{jia2014caffe}).
Similar to recent weakly-supervised learning methods~(\cite{pathak2015constrained, kolesnikov2016seed, ahn2018learning, huang2018weakly}), all networks are initialized with the weights of a pre-trained model based on the ImageNet.
All the source code and trained models will be made available online. 

{\flushleft \textbf{CAM Model.}} The CAM model is used to generate object seed regions from images based on the implementation by Ahn and Kwak~(\cite{ahn2018learning}). 
To train this CAM model, the input data is the training images and the supervisory signals are the corresponding class labels.
Similar to the CAM model by Zhou et al.~(\cite{zhou2016learning}), we use random cropping to augment data. 
For each crop, we take the class labels corresponding to the original images before cropping as supervision, and no additional supervisory signals are required.

{\flushleft \textbf{Unary Network.}}  We use the polynomial decay policy for the learning rate to train the model~(\cite{chen2018deeplab}).
The learning rate of the $k$-th iteration, $\alpha_k$,  is:
\begin{equation}
    \alpha_k = \alpha_b \times (1 - \frac{k}{K}) ^ {\tau},
\label{eq:learningrate}
\end{equation}
where the base learning rate $\alpha_b =0.001$, $\tau = 0.9$, and maximal iterations $K = 20,000$. 
The momentum parameter is set to be $0.9$.

{\flushleft \textbf{Pairwise Network.}} We use the polynomial decay policy for the learning rate as described in \eqref{eq:learningrate} to train the model, where $\alpha_b = 0.00001$, $\tau = 0.5$, $K = 20,000$, the momentum parameter is set as $0.9$.

{\flushleft \textbf{Mining Confident Regions Network.}}
We use the step learning rate decay policy.
For the $k$-th iteration, the learning rate is:
\begin{equation}
    \alpha_k = \alpha_b \times \gamma ^ {\lfloor \frac{k}{S} \rfloor},
\end{equation}
where the base learning rate $\alpha_b = 0.001$, $S = 20,000$, $\gamma=0.1$, and $\lfloor \cdot \rfloor$ is the floor function.  The momentum parameter is also set to be $0.9$.

\subsection{Performance Evaluation}
We evaluate the proposed algorithm on the PASCAL VOC 2012 dataset against the state-of-the-art weakly-supervised segmentation methods including
MIL-FCN~(\cite{pathak2014fully}), CCNN~(\cite{pathak2015constrained}), MIL-sppxl~(\cite{pinheiro2015image}), EM-Adapt~(\cite{papandreou2015weakly}), BFBP~(\cite{saleh2016built}), DCSM~(\cite{shimoda2016distinct}), AF-SS~(\cite{qi2016augmented}), AF-MCG~(\cite{qi2016augmented}), SEC~(\cite{kolesnikov2016seed}), STC~(\cite{wei2017stc}), CBTS~(\cite{roy2017combining}), AE-PSL~(\cite{wei2017object}), MCOF~(\cite{wang2018weakly}), PSA~(\cite{ahn2018learning}), DSRG~(\cite{huang2018weakly}), MDC~(\cite{wei2018revisiting}) and AISI~(\cite{fan2018associating}). 
Table~\ref{tab-res} shows the experimental results by all the evaluated methods using the VGG16~(\cite{simonyan2014very}) model as the backbone network.
The proposed algorithm achieves 62.0\% and 62.4\% on the \textit{val} and \textit{test} sets, respectively, with performance gain over the MDC~(\cite{wei2018revisiting}) method by 1.6\%. 
We note the PSA~(\cite{ahn2018learning}) model also uses affinities to refine object regions.  
However, as this method only learns affinities from coarse masks generated from CAM, the improvement by the affinity propagation is limited.
The proposed algorithm performs favorably against the PSA method by 3.6\% and 1.9\% on the \textit{val} and \textit{test} sets, respectively.
We also note that the AISI~(\cite{fan2018associating}) model recently achieves similar performance as the proposed algorithm (61.3\% on  \textit{val}, 62.1\% on  \textit{test}).
However, this method uses the $S^4$Net~(\cite{fan2017s}) to generate salient instances, which is trained with full supervision using pixel-wise annotations. 
Table 2 shows the results when the ResNet~(\cite{he2016deep}) is used as the backbone model. 
The proposed algorithm achieves performance gain over PSA~(\cite{ahn2018learning}) by 2.6\% and 1.7\% and AISI~(\cite{fan2018associating}) by 0.7\% and 0.9\% on \textit{val} and \textit{test} sets, respectively.
Figure~\ref{fig-visres} shows some segmentation results.
Overall, the segmentation results by the proposed algorithm contain fewer noisy segments. 

\begin{table}
\caption{Comparisons with the state-of-the-art weakly-supervised semantic segmentation methods on the PASCAL VOC 2012 \textit{val} set and \textit{test} set. 
All methods use the VGG16 model as the backbone network ($^\dagger$ indicates methods implicitly use full supervision).}
\label{tab-res}
\begin{tabular}{@{}p{3.6cm}ccc@{}}
\Xhline{1.0pt}
   Methods & Training Images & \textit{val} & \textit{test} \\
\hline
MIL-FCN~{\scriptsize{(ICLR'15)}}  &        10K &       25.7 &       24.9 \\

CCNN~{\scriptsize{(ICCV'15)}} &        10K &       35.3 &       35.6 \\

MIL-sppxl~{\scriptsize{(CVPR'15)}} &       700K &       36.6 &       35.8 \\

EM-Adapt~{\scriptsize{(ICCV'15)}} &        10K &       38.2 &       39.6 \\

BFBP~{\scriptsize{(ECCV'16)}} &        10K &       46.6 &       48.0 \\

DCSM~{\scriptsize{(ECCV'16)}} &        10K &       44.1 &       45.1 \\

AF-SS~{\scriptsize{(ECCV'16)}} &        10K &       52.6 &       52.7 \\

AF-MCG$^\dagger$~{\scriptsize{(ECCV'16)}} &        10K &       54.3 &       55.5 \\

SEC~{\scriptsize{(ECCV'16)}} &        10K &       50.7 &       51.7 \\

STC~{\scriptsize{(PAMI'17)}} &        50K &       49.8 &       51.2 \\

CBTS~{\scriptsize{(CVPR'17)}} &        10K &       52.8 &       53.7 \\

AE-PSL~{\scriptsize{(CVPR'17)}} &        10K &       55.0 &       55.7 \\

MCOF~{\scriptsize{(CVPR'18)}} &        10K &       56.2 &       57.6 \\

PSA~{\scriptsize{(CVPR'18)}} &        10K &       58.4 &       60.5 \\

DSRG~{\scriptsize{(CVPR'18)}} &        10K &       59.0 &       60.4 \\

MDC~{\scriptsize{(CVPR'18)}} &        10K &       60.4 &       60.8 \\

AISI$^\dagger$~{\scriptsize{(ECCV'18)}} &        10K &       61.3 &       62.1 \\
\hline
      Ours &        10K & {\bf 62.0} & {\bf 62.4} \\
\Xhline{1.0pt}
\end{tabular} 
\end{table}

\begin{table}
\caption{Evaluation results when using the ResNet as the backbone model on the PASCAL VOC 2012 dataset($^\dagger$ indicates methods implicitly use full supervision).}
\label{tab-resnet}  
\begin{tabular}{@{}p{3.6cm}ccc@{}}
\Xhline{1.0pt}
   Methods & Training Images & \textit{val} & \textit{test} \\
\hline
MCOF~{\scriptsize{(CVPR'18)}} &        10K &       60.3 &       61.2 \\

PSA~{\scriptsize{(CVPR'18)}} &        10K &       61.7 &       63.7 \\

DSRG~{\scriptsize{(CVPR'18)}} &        10K &       61.4 &       63.2 \\

AISI$^\dagger$~{\scriptsize{(ECCV'18)}} &        10K &       63.6 &       64.5 \\
\hline
 Ours &        10K & {\bf 64.3} & {\bf 65.4} \\
\Xhline{1.0pt}
\end{tabular}
\end{table}

\begin{figure*}[t]
\begin{center}
\includegraphics[width=1\linewidth,trim = 0mm 0mm 0mm 0mm, clip]{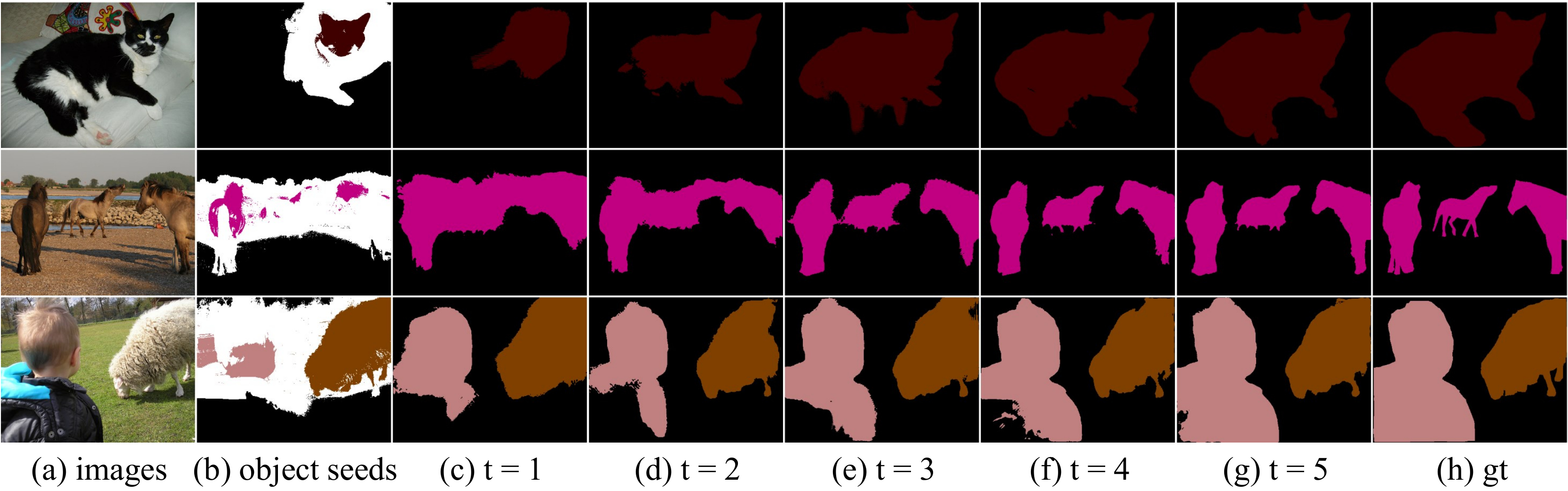}
\end{center}
\vspace{-3mm}
\caption{Visual segmentation results of each iteration of our framework on the PASCAL VOC 2012 training set. The initial object seeds are very coarse, by iteratively learning affinity, the segmentation results become better from coarse to fine. (a) images, (b) initial object seeds, (c)-(g) produced segmentation results of iterations $t=1,\ldots,5$, (h) ground truth. }
\label{fig-iterres}
\end{figure*}

\subsection{Comparison with Iterative PSA}
We note that the PSA method~(\cite{ahn2018learning}) also refines the confident regions from the CAM model  based on affinities for semantic segmentation. 
However, this approach differs significantly from our method in finding confident regions and learning affinities.
Different from the PSA model, the proposed method is optimized iteratively.
To analyze the performance of the proposed method, we design an alternative PSA approach for evaluation.
In this alternative method, confident regions are minded from the PSA model and the affinities are iteratively learned. 
We show the evaluation results\footnotemark[1] on the PASCAL VOC 2012 training set in Table~\ref{tab-psaiter}.
\footnotetext[1]{We use the code provided by the authors. The authors report results on the original training set (1464 images) of the PASCAL VOC 2012 dataset. 
Here we present results on the augmented training set (10582 images) as all models are trained on the augmented training set.}
The segmentation results of the alternative PSA model are not further refined as the number of iterations is increased.
We analyze the mined confident regions by both approaches in terms of the precision metric.
Table~\ref{tab-psaconf} shows that 
the precision of the confident regions by the alternative PSA does not increase with the number of iterations. 
This can be attributed that the PSA method determines confident object and background regions by strengthening foreground and weakening background activation maps.
We note that this approach is effective for the coarse CAM results as it can remove noisy regions. 
However, this operation also removes numerous object regions when the results are dense (e.g., large objects and complex scenes).  
Consequently, such object regions cannot be identified with more iterations.
With the learned affinities from the spatial propagation network, our method mines confident regions with a confidence network, which can remove ambiguous regions and correct noisy regions to obtain regions with higher precision.
As shown in Section~\ref{sec-formulation} and~\eqref{ytplus1}, confident regions with higher precision help the framework converge to lower energy and obtain better segmentation results, such that our approach can gradually improve with more iterations until convergence.

\begin{table} \setlength{\tabcolsep}{5pt}
\caption{Comparisons with the PSA when it is also refined iteratively. The results show the mIoU on the PASCAL VOC 2012 training set.}
\label{tab-psaiter}  
\centering
\begin{tabular}{@{}p{2.5cm}ccccc@{}}
\Xhline{1.0pt}
           &     step 1 &     step 2 &     step 3 &     step 4 &     step 5 \\
\hline
PSA &      \textbf{55.6} &       54.9 &       52.1 &      49.8     &     48.1       \\

Ours &       55.2 &      \textbf{ 59.5} &       \textbf{61.4} &       \textbf{62.7} &      \textbf{ 63.1 }\\
\Xhline{1.0pt}
\end{tabular}  
\end{table}

\begin{table}\setlength{\tabcolsep}{5pt}
\caption{Analyze the accuracy of the confident regions of the iterative PSA and ours. The results show the precision on the PASCAL VOC 2012 training set.}
\label{tab-psaconf}  
\begin{tabular}{@{}p{2.5cm}ccccc@{}}
\Xhline{1.0pt}
 & step 1 & step 2 & step 3 & step 4 & step 5 \\
\hline
PSA  &   \textbf{76.2}     &    73.7    &   71.7     &    70.2    &   68.7     \\
Ours &   73.4     &   \textbf{78.1}     &   \textbf{81.0 }    &    \textbf{81.2}    &   \textbf{81.2}     \\
\Xhline{1.0pt}
\end{tabular}
\end{table}

\begin{table}
\caption{Intermediate results of the proposed framework on the PASCAL VOC 2012 training set.}
\label{tab-interres}  
\begin{tabular}{@{}lp{2.8cm}ccc@{}}
\Xhline{1.0pt}
		 &		& 			\textit{train} & \textit{train} & \textit{val}\\
           &            &       mIoU &  Precision& mIoU \\
\hline

\multicolumn{ 1}{r}{step 0}  &      Seeds &       46.2 &       62.2 & -\\
\hline
\multicolumn{ 1}{r}{} & Unary Network &       \textbf{{\color{green}51.5}} &       72.7 \\

\multicolumn{ 1}{r}{step 1} & Mined Conf Regions &       49.8 &        \textbf{{\color{black}73.4}} \\

\multicolumn{ 1}{r}{} & Pairwise Network &      \textbf{ {\color{blue}55.2}} &       73.1  & \textbf{51.3}\\
\hline
\multicolumn{ 1}{r}{} & Unary Network &      \textbf{ {\color{green}56.6}} &       76.3 \\

\multicolumn{ 1}{r}{step 2} & Mined Conf Regions &       54.1 &      \textbf{ {\color{black}78.1}} \\

\multicolumn{ 1}{r}{} & Pairwise Network &      \textbf{ {\color{blue}59.5}} &       80.5  & \textbf{57.2}\\
\hline
\multicolumn{ 1}{r}{} & Unary Network &     \textbf{ {\color{green}59.3} }&       79.2 \\

\multicolumn{ 1}{r}{step 3} & Mined Conf Regions &       56.4 &        \textbf{{\color{black}81.0}} \\

\multicolumn{ 1}{r}{} & Pairwise Network &       \textbf{{\color{blue}61.4}} &       79.7 & \textbf{59.9}\\
\hline
\multicolumn{ 1}{r}{} & Unary Network &       \textbf{{\color{green}60.8}} &       80.7 \\

\multicolumn{ 1}{r}{step 4} & Mined Conf Regions &       56.9 &       \textbf{{\color{black}81.2}} \\

\multicolumn{ 1}{r}{} & Pairwise Network &      \textbf{ {\color{blue}62.7} }&       80.9  & \textbf{61.6}\\
\hline
\multicolumn{ 1}{r}{} & Unary Network &      \textbf{{\color{green}61.7}} &       79.7 \\

\multicolumn{ 1}{r}{step 5} & Mined Conf Regions &       57.5 &      \textbf{ {\color{black}81.2}} \\

\multicolumn{ 1}{r}{} & Pairwise Network &      \textbf{ {\color{blue}63.1}} &       80.8 & \textbf{62.0}\\

\hline
\multicolumn{ 1}{r}{} & Unary Network &      \textbf{{\color{green}61.6}} &       77.7 \\

\multicolumn{ 1}{r}{step 6} & Mined Conf Regions &       57.6 &      \textbf{ {\color{black}81.0}} \\

\multicolumn{ 1}{r}{} & Pairwise Network &      \textbf{ {\color{blue}62.8}} &       80.6 &  \textbf{61.8}\\

\hline
\multicolumn{ 1}{r}{} & Unary Network &      \textbf{{\color{green}61.8}} &       78.9 \\

\multicolumn{ 1}{r}{step 7} & Mined Conf Regions &       57.7 &      \textbf{ {\color{black}81.3}} \\

\multicolumn{ 1}{r}{} & Pairwise Network &      \textbf{ {\color{blue}63.2}} &       80.9 & \textbf{62.0}\\
\Xhline{1.0pt}
\end{tabular}  
\end{table}

\subsection{Ablation Studies}
We conduct ablation studies to analyze the contribution of each module in the proposed framework. 
All experiments are carried out on the PASCAL VOC 2012 dataset with the VGG16 model as the backbone network.

\subsubsection{Iterative Affinity Learning}
To demonstrate the effectiveness of the proposed iterative affinity learning method, we show the intermediate results on the PASCAL VOC 2012 training and validation sets in Table~\ref{tab-interres}. 
We analyze the segmentation results of the training process using the IoU and precision metric. 
As the performance of the proposed method reaches a plateau after 5 iterations,  we use the networks trained at the 5-th iteration for inference. 
With the affinity matrix being optimized in the first 5 iterations,
the performance of the unary network increases gradually from 51.5\% to 61.7\%, and that of the pairwise network increases from 55.2\% to 63.1\%. 
At each step, the mIoU of the pairwise network results is higher than that of the unary network, which demonstrates that the learned affinity matrix is effective in refining the unary segmentation network.
The main reason that the performance is increased with more iterations is that the proposed method learns robust affinities from the mined confident regions. 
Under weak supervision, we do not have integral and accurate object masks.
As we mentioned in Section~\ref{sec-pnet},
to learn robust affinities we only need some confident regions which have high precision to identify the class each region belongs to. 
As shown in Table~\ref{tab-interres}, at each step, as ambiguous regions are removed, the mined confident regions are less integral than that of the unary network (\textit{i.e.}, lower mIoU), but the precision is higher, which provides more accurate supervision for learning affinities robustly.
We also show segmentation results at each iteration in Figure~\ref{fig-iterres}. 
With more iterations, our framework gradually generates more accurate segmentation results.

\begin{table} \setlength{\tabcolsep}{3pt}
\caption{Comparisons with method that eliminates the procedure of mining confident regions. The results show the mIoU on the PASCAL VOC 2012 training set.}
\label{tab-noconf}  
\begin{tabular}{@{}p{3cm}ccccc@{}}
\Xhline{1.0pt}
           &     step 1 &     step 2 &     step 3 &     step 4 &     step 5 \\
\hline
 without mining conf. &       52.8 &       56.6 &       56.5 &       56.6     &     56.4       \\

  with mining conf. &       \textbf{55.2} &      \textbf{ 59.5} &       \textbf{61.4} &       \textbf{62.7} &      \textbf{ 63.1 }\\
\Xhline{1.0pt}
\end{tabular}  
\end{table}

\begin{table}[t] \setlength{\tabcolsep}{3pt}
\centering
\caption{Energy of each iteration without and with the procedure of mining confident regions.}
\label{tab-energy}  
\begin{tabular}{@{}p{3cm}ccccc@{}}
\Xhline{1.0pt}
           &     step 1 &     step 2 &     step 3 &     step 4 &     step 5 \\
\hline
 without mining conf. &       0.092 &       0.065 &      0.061 &       0.053     &     0.048      \\

  with mining conf. &       \textbf{0.061} &      \textbf{0.044} &       \textbf{0.042} &       \textbf{0.034} &      \textbf{ 0.029 }\\
\Xhline{1.0pt}
\end{tabular}  
\end{table}

\begin{figure*}[t]
\begin{center}
\includegraphics[width=0.98\linewidth,trim = 0mm 0mm 0mm 0mm, clip]{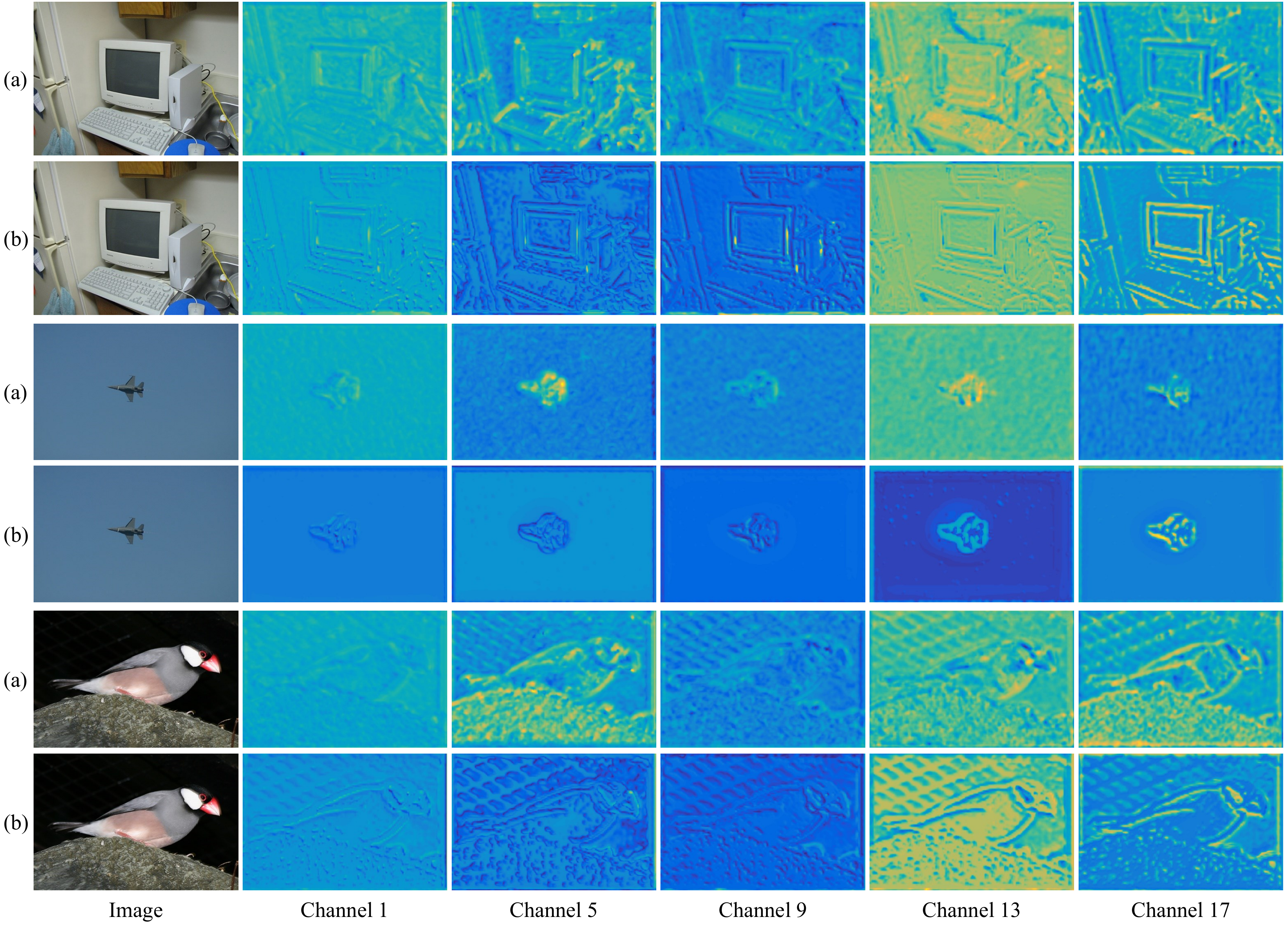}
\end{center}
\vspace{-3mm}
\caption{Visualization of the learned affinity without and with region smoothness constraint when training the pairwise affinity network. 
For each image, the first row (a) show results without the region smoothness constraint, and the second row (b) is the results with the region smoothness constraint.
With the region smoothness constraint, the learned affinity values inside object regions are smoother with more clear object boundaries. Best viewed in color.}
\label{fig-visaff}
\vspace{-2mm}
\end{figure*}

\subsubsection{Mining Regions with High Confidence Scores} \label{sec-aba-mineconf}
To validate the proposed mining method for confident regions, we compare with the alternative without this procedure. 
We show the output of the pairwise network at each iteration in Table~\ref{tab-noconf}. 
Without mining confident regions, the framework converges after 2 iterations and achieves lower segmentation performance.
With the mined confident regions for learning reliable affinity matrices, the proposed method performs better over the iterations. 
The results demonstrate the importance of the proposed mining method.

As stated in Section~\ref{sec-formulation}, by mining regions with high confidence scores, we expect they have a lower energy than original (formula (\ref{ytplus1})). 
To validate this claim, we compare the energy before and after mining the confident regions.
We randomly select 500 images as samples and compute their average energy with~(\ref{mainloss}).
Table~\ref{tab-energy} shows the intermediate results at each step.
With the proposed mining confident regions, the energy is decreased, which indicates that formula~(\ref{ytplus1}) can be satisfied.

\subsubsection{Pairwise Affinity Learning}
The pairwise affinity network aims to learn the pixel affinities to refine the segmentation results with spatial propagation.
To validate the effectiveness and necessity of learning the pairwise network, we remove it from our framework.
Table~\ref{tab-noaffinity} shows the segmentation results with and without using pixel affinities. 
Without learning the affinity network, the performance at each iteration is much lower than the proposed method.
Finally, the mIoU is lower than our method by 8.8\% on the training set and 9.2\% on the val set.
These results demonstrate the importance of learning the affinity network.
Although the proposed algorithm is able to obtain regions with high precision by mining confident regions, it misses some regions of objects, as shown in Figure~\ref{fig-mineconf}(c). 
If we directly use the mined confident regions to supervision the unary segmentation network, some segments are likely missing, and thus affect the performance.
By learning the pairwise affinity network, we can propagate the pixel labels from confident regions to regions with unknown labels. 
As such, we can achieve better object segmentation results.

\begin{table} \setlength{\tabcolsep}{2pt}
\caption{Comparisons with the alternative method without the pairwise affinity network. 
The segmentation results on the PASCAL VOC 2012 training set are presented using the mIoU.}
\label{tab-noaffinity} 
\centering 
\begin{tabular}{@{}p{3.5cm}ccccc@{}}
\Xhline{1.0pt}
           &     step 1 &     step 2 &     step 3 &     step 4 &     step 5 \\
\hline
 without learning affinity &       49.6 &     51.7 &     53.5 &     54.2  &   54.3   \\

  with learning affinity &       \textbf{55.2} &      \textbf{ 59.5} &       \textbf{61.4} &       \textbf{62.7} &      \textbf{ 63.1 }\\
\Xhline{1.0pt}
\end{tabular} 
\end{table}

\subsubsection{Region Smoothness Constraint on Affinity }
To validate the proposed region smoothness loss for training the pairwise affinity network, we show the learned pixel affinities in Figure~\ref{fig-visaff}. 
As mentioned in Section~\ref{sec-pnet}, we learn 12 affinity matrices (4 directions for 3 image channels), each affinity matrix has the same channels with the input probability maps.
For presentation clarity, here we show some channels of the first learned affinity matrix. 
The results are similar for other matrices.
With the region smoothness constraint, the learned affinity values inside object regions are smoother with more clear object boundaries.
%
For segmentation, the region smoothness constraint can improve the final results from 61.2 \% to 62.0\% on the PASCAL VOC 2012 \textit{val} set.

\begin{table}
\caption{Evaluation results on the COCO dataset.}
\label{tab-coco}  
\centering
\begin{tabular}{@{}p{6.9cm}c@{}}
\Xhline{1.0pt}
   Methods &  {mIoU} \\
\hline
SEC~{\scriptsize{(ECCV'16)}}~(\cite{kolesnikov2016seed}) &       22.4 \\

BFBP~{\scriptsize{(ECCV'16)}}~(\cite{saleh2016built}) &       20.4 \\

DSRG~{\scriptsize{(CVPR'18)}}~(\cite{huang2018weakly}) &         26.0 \\
\hline
 Ours &        {\bf 27.7} \\
\Xhline{1.0pt}
\end{tabular}
\end{table}

\subsection{Results on the COCO Dataset}
We conduct experiments on the more challenging COCO dataset, and compare with some recent methods including SEC~(\cite{kolesnikov2016seed}), BFBP~(\cite{saleh2016built}), and DSRG~(\cite{huang2018weakly}).
Table~\ref{tab-coco} shows the results on the \textit{val} set, where all methods use the VGG16 network as the backbone model. 
The proposed algorithm achieves 27.7\% on mIoU and performs favorably against the state-of-the-art methods.

\section{Conclusions}
In this paper, we propose a weakly-supervised semantic segmentation algorithm using an iterative affinity learning framework. 
Starting from the coarse annotations from the class activation maps, we exploit data redundancies in natural images to learn pixel affinities and propagate labels iteratively. 
Our framework consists of a unary segmentation network to predict the class probability map, and a pairwise affinity network to learn affinity and refine the results of the unary network. 
We propose to mine confident regions for learning the reliable affinity. 
The refined results are then considered as supervisory signals to retrain the unary network.
The procedures are conducted iteratively to learn more robust affinity and generate better segmentation progressively. 
Experimental results on both the PASCAL VOC 2012 and COCO datasets demonstrate that the proposed algorithm performs favorably against the state-of-the-art methods.

\section*{Acknowledgments}
This work is supported by National Key Basic Research Program of China (No. 2016YFB0100900),  Beijing Science and Technology Planning Project (No. Z191100007419001), National Natural Science Foundation of China (No. 61773231), and National Science Foundation (CAREER No. 1149783). 
%

%
%
%
\bibliographystyle{spbasic}
\bibliography{egbib}

\begin{thebibliography}{43}
\providecommand{\natexlab}[1]{#1}
\providecommand{\url}[1]{{#1}}
\providecommand{\urlprefix}{URL }
\expandafter\ifx\csname urlstyle\endcsname\relax
  \providecommand{\doi}[1]{DOI~\discretionary{}{}{}#1}\else
  \providecommand{\doi}{DOI~\discretionary{}{}{}\begingroup
  \urlstyle{rm}\Url}\fi
\providecommand{\eprint}[2][]{\url{#2}}

\bibitem[{Ahn and Kwak(2018)}]{ahn2018learning}
Ahn J, Kwak S (2018) Learning pixel-level semantic affinity with image-level
  supervision for weakly supervised semantic segmentation. In: Proceedings of
  IEEE Conference on Computer Vision and Pattern Recognition (CVPR), pp
  4981--4990

\bibitem[{Bearman et~al(2016)Bearman, Russakovsky, Ferrari, and
  Fei-Fei}]{bearman2016s}
Bearman A, Russakovsky O, Ferrari V, Fei-Fei L (2016) What’s the point:
  Semantic segmentation with point supervision. In: Proceedings of European
  Conference on Computer Vision (ECCV), pp 549--565

\bibitem[{Bertasius et~al(2017)Bertasius, Torresani, Stella, and
  Shi}]{bertasius2017convolutional}
Bertasius G, Torresani L, Stella XY, Shi J (2017) Convolutional random walk
  networks for semantic image segmentation. In: Proceedings of IEEE Conference
  on Computer Vision and Pattern Recognition (CVPR), pp 858--866

\bibitem[{Chen et~al(2017)Chen, Papandreou, Schroff, and
  Adam}]{chen2017rethinking}
Chen LC, Papandreou G, Schroff F, Adam H (2017) Rethinking atrous convolution
  for semantic image segmentation. arXiv preprint arXiv:170605587

\bibitem[{Chen et~al(2018)Chen, Papandreou, Kokkinos, Murphy, and
  Yuille}]{chen2018deeplab}
Chen LC, Papandreou G, Kokkinos I, Murphy K, Yuille AL (2018) Deeplab: Semantic
  image segmentation with deep convolutional nets, atrous convolution, and
  fully connected crfs. IEEE Transactions on Pattern Analysis and Machine
  Intelligence (PAMI) 40(4):834--848

\bibitem[{Dai et~al(2015)Dai, He, and Sun}]{dai2015boxsup}
Dai J, He K, Sun J (2015) Boxsup: Exploiting bounding boxes to supervise
  convolutional networks for semantic segmentation. In: Proceedings of IEEE
  International Conference on Computer Vision (ICCV), pp 1635--1643

\bibitem[{Everingham et~al(2010)Everingham, Van~Gool, Williams, Winn, and
  Zisserman}]{everingham2010pascal}
Everingham M, Van~Gool L, Williams CK, Winn J, Zisserman A (2010) The pascal
  visual object classes (voc) challenge. International Journal of Computer
  Vision (IJCV) 88(2):303--338

\bibitem[{Fan et~al(2018)Fan, Hou, Cheng, Yu, Martin, and
  Hu}]{fan2018associating}
Fan R, Hou Q, Cheng MM, Yu G, Martin RR, Hu SM (2018) Associating inter-image
  salient instances for weakly supervised semantic segmentation. In:
  Proceedings of European Conference on Computer Vision (ECCV), pp 367--383

\bibitem[{Fan et~al(2019)Fan, Cheng, Hou, Mu, Wang, and Hu}]{fan2017s}
Fan R, Cheng MM, Hou Q, Mu TJ, Wang J, Hu SM (2019) S4net: Single stage
  salient-instance segmentation. In: Proceedings of IEEE Conference on Computer
  Vision and Pattern Recognition (CVPR), pp 6103--6112

\bibitem[{Felzenszwalb and Huttenlocher(2004)}]{felzenszwalb2004efficient}
Felzenszwalb PF, Huttenlocher DP (2004) Efficient graph-based image
  segmentation. International Journal of Computer Vision (IJCV) 59(2):167--181

\bibitem[{Hagen and Kahng(1992)}]{hagen1992new}
Hagen L, Kahng AB (1992) New spectral methods for ratio cut partitioning and
  clustering. IEEE Transactions on Computer-Aided Design of Integrated Circuits
  and Systems pp 1074--1085

\bibitem[{Hariharan et~al(2011)Hariharan, Arbel{\'a}ez, Bourdev, Maji, and
  Malik}]{hariharan2011semantic}
Hariharan B, Arbel{\'a}ez P, Bourdev L, Maji S, Malik J (2011) Semantic
  contours from inverse detectors. In: Proceedings of IEEE International
  Conference on Computer Vision (ICCV), pp 991--998

\bibitem[{He et~al(2016)He, Zhang, Ren, and Sun}]{he2016deep}
He K, Zhang X, Ren S, Sun J (2016) Deep residual learning for image
  recognition. In: Proceedings of IEEE Conference on Computer Vision and
  Pattern Recognition (CVPR), pp 770--778

\bibitem[{Huang et~al(2018)Huang, Wang, Wang, Liu, and Wang}]{huang2018weakly}
Huang Z, Wang X, Wang J, Liu W, Wang J (2018) Weakly-supervised semantic
  segmentation network with deep seeded region growing. In: Proceedings of IEEE
  Conference on Computer Vision and Pattern Recognition (CVPR), pp 7014--7023

\bibitem[{Jia et~al(2014)Jia, Shelhamer, Donahue, Karayev, Long, Girshick,
  Guadarrama, and Darrell}]{jia2014caffe}
Jia Y, Shelhamer E, Donahue J, Karayev S, Long J, Girshick R, Guadarrama S,
  Darrell T (2014) Caffe: Convolutional architecture for fast feature
  embedding. In: Proceedings of ACM international conference on Multimedia (ACM
  MM), pp 675--678

\bibitem[{Kersten(1987)}]{kersten1987predictability}
Kersten D (1987) Predictability and redundancy of natural images. JOSA A
  4(12):2395--2400

\bibitem[{Khoreva et~al(2017)Khoreva, Benenson, Hosang, Hein, and
  Schiele}]{khoreva2017simple}
Khoreva A, Benenson R, Hosang J, Hein M, Schiele B (2017) Simple does it:
  Weakly supervised instance and semantic segmentation. In: Proceedings of IEEE
  Conference on Computer Vision and Pattern Recognition (CVPR), pp 876--885

\bibitem[{Kolesnikov and Lampert(2016)}]{kolesnikov2016seed}
Kolesnikov A, Lampert CH (2016) Seed, expand and constrain: Three principles
  for weakly-supervised image segmentation. In: Proceedings of European
  Conference on Computer Vision (ECCV), pp 695--711

\bibitem[{Levin et~al(2008)Levin, Lischinski, and Weiss}]{Levin2008ACS}
Levin A, Lischinski D, Weiss Y (2008) A closed-form solution to natural image
  matting. IEEE Transactions on Pattern Analysis and Machine Intelligence
  (PAMI) 30:228--242

\bibitem[{Lin et~al(2016)Lin, Dai, Jia, He, and Sun}]{lin2016scribblesup}
Lin D, Dai J, Jia J, He K, Sun J (2016) Scribblesup: Scribble-supervised
  convolutional networks for semantic segmentation. In: Proceedings of IEEE
  Conference on Computer Vision and Pattern Recognition (CVPR), pp 3159--3167

\bibitem[{Lin et~al(2014)Lin, Maire, Belongie, Hays, Perona, Ramanan,
  Doll{\'a}r, and Zitnick}]{lin2014microsoft}
Lin TY, Maire M, Belongie S, Hays J, Perona P, Ramanan D, Doll{\'a}r P, Zitnick
  CL (2014) Microsoft {COCO}: Common objects in context. In: Proceedings of
  European Conference on Computer Vision (ECCV), pp 740--755

\bibitem[{Liu et~al(2017)Liu, De~Mello, Gu, Zhong, Yang, and
  Kautz}]{liu2017learning}
Liu S, De~Mello S, Gu J, Zhong G, Yang MH, Kautz J (2017) Learning affinity via
  spatial propagation networks. In: Proceedings of Annual Conference on Neural
  Information Processing Systems (NeurIPS), pp 1520--1530

\bibitem[{Long et~al(2015)Long, Shelhamer, and Darrell}]{long2015fully}
Long J, Shelhamer E, Darrell T (2015) Fully convolutional networks for semantic
  segmentation. In: Proceedings of IEEE Conference on Computer Vision and
  Pattern Recognition (CVPR), pp 3431--3440

\bibitem[{Maire et~al(2016)Maire, Narihira, and Yu}]{maire2016affinity}
Maire M, Narihira T, Yu SX (2016) Affinity {CNN}: Learning pixel-centric
  pairwise relations for figure/ground embedding. In: Proceedings of IEEE
  Conference on Computer Vision and Pattern Recognition (CVPR), pp 174--182

\bibitem[{Papandreou et~al(2015)Papandreou, Chen, Murphy, and
  Yuille}]{papandreou2015weakly}
Papandreou G, Chen LC, Murphy KP, Yuille AL (2015) Weakly-and semi-supervised
  learning of a deep convolutional network for semantic image segmentation. In:
  Proceedings of IEEE International Conference on Computer Vision (ICCV), pp
  1742--1750

\bibitem[{Pathak et~al(2014)Pathak, Shelhamer, Long, and
  Darrell}]{pathak2014fully}
Pathak D, Shelhamer E, Long J, Darrell T (2014) Fully convolutional multi-class
  multiple instance learning. arXiv preprint arXiv:14127144

\bibitem[{Pathak et~al(2015)Pathak, Krahenbuhl, and
  Darrell}]{pathak2015constrained}
Pathak D, Krahenbuhl P, Darrell T (2015) Constrained convolutional neural
  networks for weakly supervised segmentation. In: Proceedings of IEEE
  International Conference on Computer Vision (ICCV), pp 1796--1804

\bibitem[{Pinheiro and Collobert(2015)}]{pinheiro2015image}
Pinheiro PO, Collobert R (2015) From image-level to pixel-level labeling with
  convolutional networks. In: Proceedings of IEEE Conference on Computer Vision
  and Pattern Recognition (CVPR), pp 1713--1721

\bibitem[{Qi et~al(2016)Qi, Liu, Shi, Zhao, and Jia}]{qi2016augmented}
Qi X, Liu Z, Shi J, Zhao H, Jia J (2016) Augmented feedback in semantic
  segmentation under image level supervision. In: Proceedings of European
  Conference on Computer Vision (ECCV), pp 90--105

\bibitem[{Roy and Todorovic(2017)}]{roy2017combining}
Roy A, Todorovic S (2017) Combining bottom-up, top-down, and smoothness cues
  for weakly supervised image segmentation. In: Proceedings of IEEE Conference
  on Computer Vision and Pattern Recognition (CVPR), pp 3529--3538

\bibitem[{Saleh et~al(2016)Saleh, Aliakbarian, Salzmann, Petersson, Gould, and
  Alvarez}]{saleh2016built}
Saleh F, Aliakbarian MS, Salzmann M, Petersson L, Gould S, Alvarez JM (2016)
  Built-in foreground/background prior for weakly-supervised semantic
  segmentation. In: Proceedings of European Conference on Computer Vision
  (ECCV), pp 413--432

\bibitem[{Shi and Malik(2000)}]{shi2000normalized}
Shi J, Malik J (2000) Normalized cuts and image segmentation. IEEE Transactions
  on Pattern Analysis and Machine Intelligence (PAMI) 22(8):888--905

\bibitem[{Shimoda and Yanai(2016)}]{shimoda2016distinct}
Shimoda W, Yanai K (2016) Distinct class-specific saliency maps for weakly
  supervised semantic segmentation. In: Proceedings of European Conference on
  Computer Vision (ECCV), pp 218--234

\bibitem[{Simonyan and Zisserman(2014)}]{simonyan2014very}
Simonyan K, Zisserman A (2014) Very deep convolutional networks for large-scale
  image recognition. arXiv preprint arXiv:14091556

\bibitem[{Wang et~al(2018{\natexlab{a}})Wang, Ma, Chen, and You}]{wang2018edge}
Wang X, Ma H, Chen X, You S (2018{\natexlab{a}}) Edge preserving and
  multi-scale contextual neural network for salient object detection. IEEE
  Transactions on Image Processing (TIP) 27(1):121--134

\bibitem[{Wang et~al(2018{\natexlab{b}})Wang, You, Li, and Ma}]{wang2018weakly}
Wang X, You S, Li X, Ma H (2018{\natexlab{b}}) Weakly-supervised semantic
  segmentation by iteratively mining common object features. In: Proceedings of
  IEEE Conference on Computer Vision and Pattern Recognition (CVPR), pp
  1354--1362

\bibitem[{Wei et~al(2017{\natexlab{a}})Wei, Feng, Liang, Cheng, Zhao, and
  Yan}]{wei2017object}
Wei Y, Feng J, Liang X, Cheng MM, Zhao Y, Yan S (2017{\natexlab{a}}) Object
  region mining with adversarial erasing: A simple classification to semantic
  segmentation approach. In: Proceedings of IEEE Conference on Computer Vision
  and Pattern Recognition (CVPR), pp 1568--1576

\bibitem[{Wei et~al(2017{\natexlab{b}})Wei, Liang, Chen, Shen, Cheng, Feng,
  Zhao, and Yan}]{wei2017stc}
Wei Y, Liang X, Chen Y, Shen X, Cheng MM, Feng J, Zhao Y, Yan S
  (2017{\natexlab{b}}) {STC}: A simple to complex framework for
  weakly-supervised semantic segmentation. IEEE Transactions on Pattern
  Analysis and Machine Intelligence (PAMI) 39(11):2314--2320

\bibitem[{Wei et~al(2018)Wei, Xiao, Shi, Jie, Feng, and
  Huang}]{wei2018revisiting}
Wei Y, Xiao H, Shi H, Jie Z, Feng J, Huang TS (2018) Revisiting dilated
  convolution: A simple approach for weakly-and semi-supervised semantic
  segmentation. In: Proceedings of IEEE Conference on Computer Vision and
  Pattern Recognition (CVPR), pp 7268--7277

\bibitem[{Wei et~al(1989)Wei, Cheng et~al}]{wei1989towards}
Wei YC, Cheng CK, et~al (1989) Towards efficient hierarchical designs by ratio
  cut partitioning. In: IEEE International Conference on Computer-Aided Design,
  pp 298--301

\bibitem[{Zhao et~al(2017)Zhao, Shi, Qi, Wang, and Jia}]{zhao2017pyramid}
Zhao H, Shi J, Qi X, Wang X, Jia J (2017) Pyramid scene parsing network. In:
  Proceedings of IEEE Conference on Computer Vision and Pattern Recognition
  (CVPR), pp 2881--2890

\bibitem[{Zhou et~al(2016)Zhou, Khosla, Lapedriza, Oliva, and
  Torralba}]{zhou2016learning}
Zhou B, Khosla A, Lapedriza A, Oliva A, Torralba A (2016) Learning deep
  features for discriminative localization. In: Proceedings of IEEE Conference
  on Computer Vision and Pattern Recognition (CVPR), pp 2921--2929

\bibitem[{Zhou et~al(2019)Zhou, Zhao, Puig, Xiao, Fidler, Barriuso, and
  Torralba}]{zhou2019semantic}
Zhou B, Zhao H, Puig X, Xiao T, Fidler S, Barriuso A, Torralba A (2019)
  Semantic understanding of scenes through the ade20k dataset. International
  Journal of Computer Vision (IJCV) 127(3):302--321

\end{thebibliography}
\end{sloppypar}
\end{document}